\documentclass{article}

\usepackage{arxiv}

\usepackage[utf8]{inputenc}
\usepackage[T1]{fontenc}
\usepackage{hyperref}
\usepackage{url}
\usepackage{booktabs}
\usepackage{amsmath}
\usepackage{amssymb}
\usepackage{amsfonts}
\usepackage{nicefrac}
\usepackage{microtype}
\usepackage{graphicx}
\usepackage{tabularx}
\usepackage{natbib}
\usepackage{doi}
\usepackage{multirow}
\title{ED-DiT: Physics-Guided Diffusion Pretraining for Transferable Molecular Representations from Electron Density}

\author{
\begin{tabular}{c}
Shuang Liang$^{1}$ \quad
Haocheng Wang$^{2}$ \quad
Jiayi Song$^{1}$ \\
Shuquan Ye$^{2}$ \quad
Ben Fei$^{3,*}$ \\[2mm]
$^{1}$Fudan University, Shanghai, China \\
$^{2}$Shenzhen Loop Area Institute, Shenzhen, China \\
$^{3}$The Chinese University of Hong Kong, Hong Kong SAR, China \\[1mm]
{\small\texttt{23300110095@m.fudan.edu}
\quad \texttt{jiayisong@tamu.edu}} \\
{\small\texttt{haochengwa@gmail.com}
\quad \texttt{shuquanye@slai.edu.cn}} \\
{\small\texttt{benfei@cuhk.edu.hk}} \\[1mm]
{\small $^{*}$Corresponding author}
\end{tabular}
}
\date{}

\renewcommand{\shorttitle}{ED-DiT}

\hypersetup{
  pdftitle={ED-DiT: Physics-Guided Diffusion Pretraining for Transferable Molecular Representations from Electron Density},
  pdfsubject={Machine Learning; Computational Chemistry; Electron Density},
    pdfauthor={Liang Shuang, Haocheng Wang, Jiayi Song, Shuquan Ye, Ben Fei},
  pdfkeywords={electron density, molecular representation learning, diffusion transformer, self-supervised pretraining, physics-guided learning},
}

\begin{document}

\maketitle

\begin{abstract} 
Pretraining has shown strong potential for learning transferable representations, yet it remains underexplored for electron-density-based molecular learning. Electron density provides a continuous three-dimensional description of molecular electronic structure, capturing both local spatial patterns and global physical quantities. This raises a key question: can electron-density fields be used for self-supervised pretraining to learn a shared representation that transfers across diverse electronic-structure-related tasks? We propose \textbf{ED-DiT}, a physics-guided Diffusion Transformer for self-supervised pretraining on electron-density point clouds. ED-DiT learns reusable representations by reconstructing corrupted and partially masked log-density fields across diffusion noise levels. An electron-number consistency constraint is further introduced to preserve the total electronic mass. The pretrained encoder can be adapted to property prediction, open-/closed-shell classification, molecule--electron-density retrieval, and molecule-conditioned electron-density prediction. Experiments on six EDBench tasks show that ED-DiT consistently outperforms the same architecture trained from scratch, especially under limited supervision. For molecule-conditioned electron-density prediction, it reduces RMSE from 2.2474 to 1.3753 and surpasses the available baseline. With only 10\% labels, it improves orbital energy prediction RMSE from 0.0293 to 0.0138. These results demonstrate the effectiveness of physics-guided electron-density pretraining for learning transferable molecular representations. \end{abstract}

{%
\section{Introduction}
Pretraining has become a central strategy for learning reusable representations from large-scale, task-agnostic data, enabling models to adapt to diverse downstream tasks with limited supervision. In computer vision, reconstruction, contrastive learning, and generative objectives have demonstrated the effectiveness of this pretraining-and-transfer paradigm across a wide range of applications~\citep{he2022masked, chen2020simple, hudson2024soda}. A similar need arises in molecular machine learning, where diverse quantum-property and electronic-structure tasks are often addressed with task-specific supervision and independently trained models. 
Electron density offers a continuous, three-dimensional, and physically meaningful description of the spatial distribution of electrons, encoding information such as the total electron number, spatial density distribution, and local variations that are closely related to molecular electronic structure~\citep{hohenberg1964inhomogeneous, xiang2025edbench}. This motivates the central question of this work: can three-dimensional electron-density data be leveraged through self-supervised pretraining to learn molecular representations that transfer across diverse electronic-structure-related tasks?

Existing molecular representation learning methods have demonstrated the effectiveness of pretraining on modalities such as SMILES strings, molecular graphs, atomic coordinates, and generic three-dimensional structures~\citep{ross2022large, feng2024unicorn,ji2024exploring}. The resulting representations have been successfully transferred across diverse downstream applications, demonstrating the feasibility of learning reusable molecular representations beyond individual tasks. 
Electron-density-based learning, however, has so far been studied mainly in task-specific settings, including property prediction, classification, cross-modal retrieval, and density-field prediction~\citep{parrilla2024electron,li2025electron}. In these studies, electron density is typically treated as either an input modality or a prediction target, with separate models developed for individual tasks. Large-scale resources such as EDBench now provide unified electron-density data and diverse downstream tasks, creating the conditions for systematically evaluating whether a shared representation can transfer across tasks~\citep{xiang2025edbench}. 
Recent work has also explored self-supervised diffusion for electron-aware molecular representation learning ~\citep{na2025self}. Nevertheless, self-supervised pretraining directly on explicit three-dimensional electron-density fields to learn a shared encoder transferable across heterogeneous electron-density-centric tasks remains underexplored.
Therefore, learning transferable representations from electron density presents challenges beyond conventional geometric pretraining. Although electron-density fields can be discretized into point-based representations, they are fundamentally three-dimensional physical fields whose integral corresponds to the total electron number, a basic physical quantity that generic geometric objectives do not explicitly preserve. Generic masked modeling or reconstruction objectives may recover local density patterns while violating this global physical constraint. Consequently, such objectives may produce spatially plausible reconstructions while the learned representations may fail to preserve physical consistency, which may in turn hinder their transfer to downstream molecular tasks. This calls for electron-density pretraining objectives that jointly learn transferable representations and explicitly preserve electron-number consistency.

To jointly learn transferable representations and preserve physical consistency, we propose \textbf{ED-DiT}, a physics-guided Diffusion Transformer framework ~\citep{peebles2023scalable} for self-supervised pretraining on three-dimensional electron densities. ED-DiT represents electron density as a point-based density field and uses a Diffusion Transformer to encode its spatial coordinates and density values. During pretraining, diffusion noise of varying intensities is applied to randomly masked log-density tokens, and the model learns to predict the injected noise and recover the corresponding density values from the uncorrupted context. Learning across multiple noise levels encourages the encoder to capture robust and reusable representations of electron-density fields. In addition to masked diffusion denoising, ED-DiT incorporates an electron-number consistency constraint that explicitly preserves the total electronic mass of the reconstructed density field. After pretraining, the shared ED-DiT encoder is transferred to six EDBench downstream tasks~\citep{xiang2025edbench}: orbital energy prediction, energy correction, multipole moment prediction, open-/closed-shell classification, molecule–electron-density cross-modal retrieval, and molecule-conditioned electron-density prediction. Experiments show that ED-DiT outperforms the same architecture trained from scratch across most downstream settings, with particularly strong gains under limited supervision; under the 10\% setting, it improves RMSE on MM from 0.899 to 0.381 and reduces RMSE on OE by 52.97\%. Ablation and diagnostic analyses further show that the electron-number constraint substantially improves the physical consistency of reconstructed density fields and can also benefit downstream transfer, supporting physics-guided electron-density pretraining as a practical approach to data-efficient molecular representation learning.

Our main contributions are summarized as follows:
\begin{itemize}
    \item We formulate self-supervised pretraining on three-dimensional electron density as a transferable molecular representation learning problem, and systematically study whether a shared electron-density encoder can support diverse downstream tasks beyond task-specific modeling.
    \item We propose \textbf{ED-DiT}, a physics-guided Diffusion Transformer that learns from electron-density fields through masked multi-noise-level denoising and explicitly preserves the total electron number of reconstructed density fields through an electron-number consistency constraint.
    \item We conduct comprehensive evaluations on six EDBench tasks spanning property regression, classification, cross-modal retrieval, and density-field prediction. The results demonstrate consistent advantages over training from scratch, particularly under limited supervision.
\end{itemize}
}

\section{Related Work}
\paragraph{Molecular tasks and molecular foundation models.}
Molecular machine learning covers diverse tasks, such as property prediction, molecular generation, and cross-modal retrieval. 
For task-specific molecular learning, MoleculeNet provides standardized benchmarks for property prediction under limited supervision ~\citep{wu2018moleculenet}, while geometric architectures such as SchNet, DimeNet, PaiNN, and GemNet model atomic interactions and symmetry-aware features ~\citep{schutt2017schnet,gasteiger2020dimenet,schutt2021painn,gasteiger2021gemnet}.
Molecular foundation models instead aim to learn transferable representations through large-scale pretraining~\citep{ji2024exploring}. ChemBERTa and MolFormer learn from SMILES strings ~\citep{chithrananda2020chemberta,ross2022large}, GROVER and MolCLR pretrain on molecular graphs ~\citep{rong2020grover,wang2021molclr}, and other methods incorporate three-dimensional conformations or align two- and three-dimensional molecular views ~\citep{liu2021graphmvp,stark2021infomax,zhu2022unified}. Multimodal approaches further connect molecular representations with natural language for retrieval and generation ~\citep{liu2022moleculestm,pei20253d}. 
Beyond molecule-specific models, SciReasoner unifies natural language with heterogeneous scientific sequences in a cross-disciplinary foundation model ~\citep{wang2025scireasoner}. Nevertheless, these approaches primarily learn from molecular strings, graphs, coordinates, or other symbolic and structural representations, rather than directly pretraining on electron-density fields.

\paragraph{Electron-density-based molecular learning.}
Electron density provides a spatial description of molecular electronic structure and has been widely studied as a prediction target. Neural methods accelerate electron-density estimation from molecular structures \citep{sinitskiy2018electron,fu2024recipe,elsborg2026electra}, with subsequent approaches incorporating atom--query-point message passing and geometric equivariance to improve density prediction \citep{jorgensen2020deepdft,jorgensen2021equivariant}. 
More recently, EDBench introduced a unified benchmark covering property prediction, cross-modal retrieval, and molecule-conditioned density prediction \citep{xiang2025edbench}. However, existing methods remain largely task-specific, treating electron density as either an input or a prediction target~\citep{parrilla2024electron,li2025electron}, while self-supervised pretraining of a transferable electron-density encoder remains underexplored.

\paragraph{3D self-supervised learning and physical constraints.}
Self-supervised pretraining for 3D point clouds can be broadly grouped into discriminative and generative paradigms \citep{fei2023self}. Discriminative approaches, such as PointContrast, learn invariant representations through contrastive objectives \citep{xie2020pointcontrast}, whereas generative approaches, including Point-BERT and Point-MAE, learn by predicting or reconstructing masked geometric content \citep{yu2021pointbert,pang2022pointmae}. 
Unlike ordinary geometric point clouds, electron density is a physical field whose integral corresponds to the total electron number, a quantity not explicitly preserved by generic reconstruction objectives. ED-DiT therefore extends generative 3D pretraining with masked diffusion denoising and an electron-number consistency constraint.

\begin{figure*}[t]
    \centering
    \includegraphics[width=\textwidth]{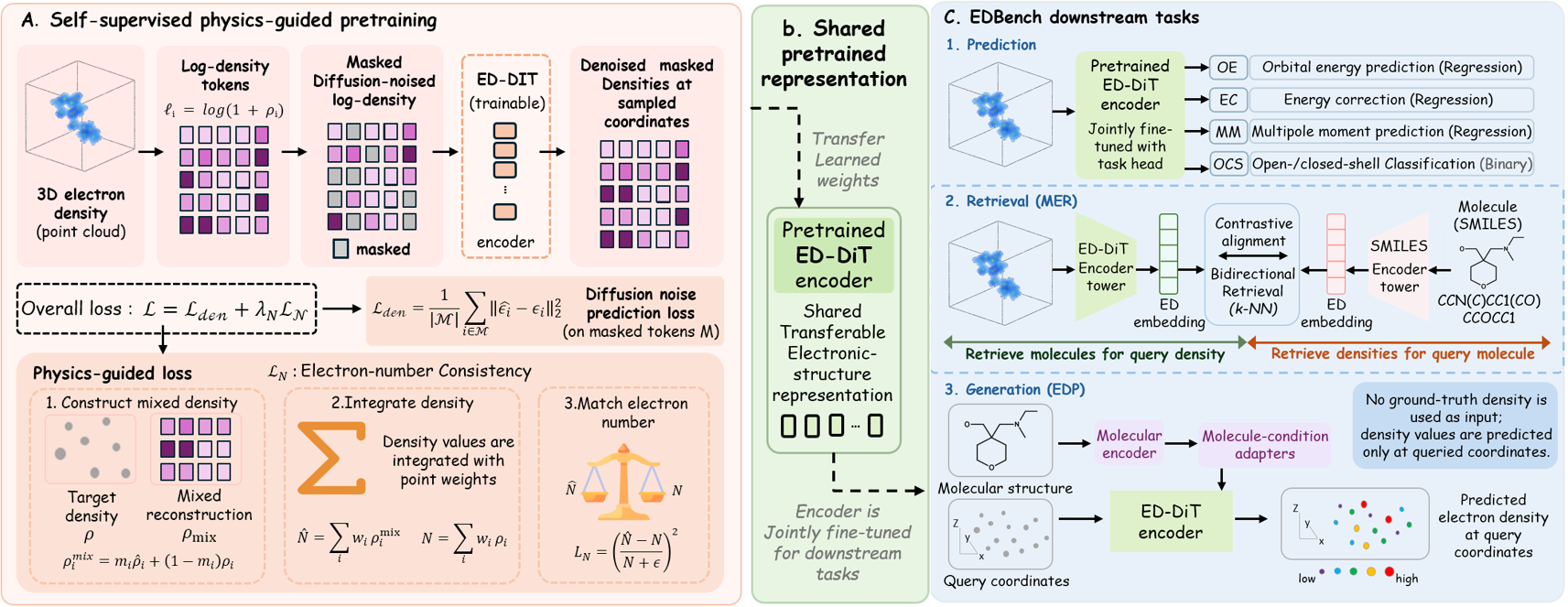}
    \caption{Overview of ED-DiT. 
(a) ED-DiT performs self-supervised physics-guided pretraining on 3D electron-density point clouds by predicting diffusion noise on masked log-density tokens. The pretraining objective combines masked denoising with an electron-number constraint. 
(b) The pretrained encoder learns a shared transferable electronic-structure representation. 
(c) The representation is adapted to EDBench downstream tasks, including property prediction, open-/closed-shell classification, molecule-electron-density retrieval, and molecule-conditioned electron-density prediction.}
    \label{fig:overview}
\end{figure*}
\section{Method}

\subsection{Overview}

ED-DiT learns a transferable electron-density representation through self-supervised pretraining on molecular electron-density fields. As illustrated in Figure~\ref{fig:overview}, the framework consists of three stages. First, electron-density fields are converted into spatial point representations that encode both molecular geometry and electronic information. Second, a Diffusion Transformer backbone is pretrained to reconstruct corrupted electron-density fields, allowing the model to learn reusable electronic-structure representations without downstream labels. Physics-guided constraints are incorporated to improve the physical consistency of the learned representation. Finally, the pretrained backbone is adapted to diverse EDBench tasks~\citep{xiang2025edbench}, including molecular property prediction, open-/closed-shell classification, molecule--electron-density retrieval, and molecule-conditioned electron-density prediction.

\subsection{Electron-Density Backbone}

The backbone is designed to convert irregular three-dimensional electron-density samples into normalized point tokens, enabling ED-DiT to learn both local density patterns and global electronic-structure representations.
\paragraph{Electron-Density Point Representation.}
We represent each molecule as a three-dimensional ED point field:
\begin{equation}
    \mathcal{P}=\{(\mathbf r_i,\rho_i,w_i)\}_{i=1}^{N},
\end{equation}
where $\mathbf r_i\in\mathbb{R}^{3}$ is the spatial coordinate, $\rho_i\geq 0$ is the electron-density value, and $w_i$ is the sample weight
for numerical integration. In practice, we use uniform weights 
$w_i=1/N_{\mathrm{valid}}$ so that ED integrals can be approximated by 
$\int \rho(\mathbf r)d\mathbf r \approx \sum_i w_i\rho_i$.  Since electron-density values have a large dynamic range, we transform them into log-density features as $d_i=\log(1+\rho_i)$. 
For each molecule, coordinates are normalized by centering valid points and scaling by the maximum radius:
\begin{equation}
    \mathbf c_r=\frac{1}{\sum_i v_i}\sum_i v_i\mathbf r_i,\quad
    R=\max_i v_i\|\mathbf r_i-\mathbf c_r\|_2,
\end{equation}
\begin{equation}
    \bar{\mathbf r}_i=
    \frac{\mathbf r_i-\mathbf c_r}{R+\epsilon},
\end{equation}
where $v_i\in\{0,1\}$ indicates whether point $i$ is valid, $\mathbf c_r$ is the center of valid points, $R$ is the maximum radius, $\epsilon$ is a small constant for numerical stability, and $\bar{\mathbf r}_i$ is the normalized coordinate. The model therefore operates on normalized coordinates and log-density features.

\paragraph{Diffusion Transformer Encoder.}
Given normalized coordinates and an input density feature $d_i^{in}$, ED-DiT constructs each point token as
\begin{equation}
    \mathbf z_i^{(0)}
    =
    \phi_{in}([\bar{\mathbf r}_i;d_i^{in}])
    +
    \phi_{pos}(\bar{\mathbf r}_i),
\end{equation}
where $\phi_{in}$ is an input MLP and $\phi_{pos}$ is a coordinate positional MLP. For diffusion pretraining, the timestep $t$ is embedded as
\begin{equation}
    \mathbf c_t=\phi_t(\mathrm{PE}(t)),
\end{equation}
where $\mathrm{PE}(\cdot)$ denotes sinusoidal timestep embedding.

The point tokens are processed by a stack of DiT blocks:
\begin{equation}
    \mathbf Z^{(\ell+1)}
    =
    \mathrm{DiTBlock}_{\ell}
    (\mathbf Z^{(\ell)},\mathbf c_t,\mathbf v),
    \quad \ell=0,\ldots,L-1.
\end{equation}
where $\mathbf Z^{(\ell)}=\{\mathbf z_i^{(\ell)}\}_{i=1}^{N}$ denotes the set of point-token representations at layer $\ell$, and $\mathbf v=(v_1,\ldots,v_N)$ is the valid-point mask. 
Each block applies self-attention over valid ED points and uses adaptive layer normalization modulated by the timestep embedding. The encoder produces local point features and a global ED representation:
\begin{equation}
    \mathbf h_i=\mathbf z_i^{(L)},\quad
    \mathbf h_{\mathcal P}
    =
    \frac{\sum_i v_i\mathbf h_i}{\sum_i v_i}.
\end{equation}
Local point features are used for denoising during pretraining, while the global representation is used for downstream prediction and retrieval tasks.

\subsection{Self-Supervised ED Pretraining}

\paragraph{Masked Diffusion Denoising.}
This objective encourages the encoder to learn transferable electron-density representations by recovering masked noisy density values from their surrounding context~\citep{ho2020denoising}. 

ED-DiT is pretrained by reconstructing corrupted ED fields without using downstream labels. Given the log-density value $d_i$, we sample a diffusion timestep $t$ and Gaussian noise $\epsilon_i\sim\mathcal{N}(0,1)$. The noisy log-density is
\begin{equation}
    \tilde d_{i,t}
    =
    \sqrt{\bar{\alpha}_t}d_i
    +
    \sqrt{1-\bar{\alpha}_t}\epsilon_i,
\end{equation}
where $\bar{\alpha}_t=\prod_{s=1}^{t}\alpha_s$.
To make denoising contextual, we independently mask each valid ED point with probability 0.5. Let $m_i\in\{0,1\}$ indicate whether point $i$ is masked. The density feature given to the encoder is
\begin{equation}
    d_i^{in}
    =
    m_i\tilde d_{i,t}
    +
    (1-m_i)d_i.
\end{equation}
Thus, masked points receive diffusion-noised log-density values, while unmasked points provide clean context.
The encoder predicts the injected noise:
\begin{equation}
    \hat{\epsilon}_i
    =
    q_{\theta}(\bar{\mathbf r}_i,d_i^{in},t),
\end{equation}
and the denoising objective is computed only on masked points:
\begin{equation}
    \mathcal{L}_{den}
    =
    \frac{1}{|\mathcal M|}
    \sum_{i\in\mathcal M}
    \|\hat{\epsilon}_i-\epsilon_i\|_2^2,
\end{equation}
where $\mathcal M=\{i:m_i=1\}$.
\paragraph{Electron-Number Consistency.}

Masked denoising learns local density patterns, but it does not explicitly
encourage global physical consistency. We therefore regularize the reconstructed
density field with a point-sampled proxy for electron-number consistency.
The predicted noise is converted back to a denoised log-density:
\begin{equation}
    \hat d_i
    =
    \frac{
    \tilde d_{i,t}
    -
    \sqrt{1-\bar{\alpha}_t}\hat{\epsilon}_i
    }{
    \sqrt{\bar{\alpha}_t}
    },
\end{equation}
and the reconstructed density is $\hat{\rho}_i=\exp(\hat d_i)-1$.
In practice, $\hat d_i$ is clipped to $[0,d_{\max}]$ for numerical stability,
where $d_{\max}=\log(1+4\max_j \rho_j)$ is computed for each molecule, and
$\hat{\rho}_i$ is clamped to be non-negative. Physical consistency is applied
to a mixed density field:
\begin{equation}
    \rho_i^{mix}
    =
    m_i\hat{\rho}_i+(1-m_i)\rho_i,
\end{equation}
where the reconstructed density is used on masked points, and the ground-truth
density is retained on unmasked points. We define the point-sampled
electron-number proxy as
\begin{equation}
    N_p(\rho)=\sum_i w_i\rho_i,
\end{equation}
where $w_i$ is the sampling weight of point $i$. In our implementation,
$w_i=1/N_{\mathrm{valid}}$, so $N_p(\rho)$ preserves the global density
magnitude over sampled ED points and serves as a proxy for the continuous
electron number. The electron-number consistency loss is
\begin{equation}
    \mathcal L_N
    =
    \left(
    \frac{
    N_p(\rho^{mix})-N_p(\rho)
    }{
    N_p(\rho)+\epsilon
    }
    \right)^2.
\end{equation}
\paragraph{Pretraining Objective.}

The final ED-DiT pretraining objective combines local denoising with global electron-number consistency:
\begin{equation}
    \mathcal L_{pre}
    =
    \mathcal L_{den}
    +
    \lambda_N\mathcal L_N.
\end{equation}
This objective encourages the backbone to learn both spatial density patterns and physically meaningful global ED structure.

\subsection{Downstream Adaptation}

After pretraining, the ED-DiT encoder is transferred to EDBench downstream tasks. The same pretrained backbone is reused across tasks, while task-specific heads or adapters are learned during supervised fine-tuning.

\paragraph{Property Prediction and Classification.}

For property regression tasks, including orbital energy prediction, energy correction, and multipole moment prediction, the global representation is fed into a regression head:
\begin{equation}
    \hat{\mathbf y}
    =
    g_{\phi}^{reg}(\mathbf h_{\mathcal P}).
\end{equation}
where $g_{\phi}^{reg}$ denotes the task-specific regression head with parameters $\phi$.
For open-/closed-shell classification, a classification head predicts the binary state:
    $p(y=1|\mathcal P)
    =
    \sigma(g_{\phi}^{cls}(\mathbf h_{\mathcal P}))$,
where $g_{\phi}^{cls}$ denotes the task-specific classification head with parameters $\phi$, and $\sigma(\cdot)$ is the sigmoid function.
In both cases, the pretrained encoder and the task-specific head are jointly optimized.

\paragraph{Molecule Electron Density Retrieval.}

For molecule electron density retrieval, ED-DiT serves as the electron-density encoder. A molecule encoder processes the paired molecular input, and both modalities are projected into a shared embedding space:
\begin{equation}
    \mathbf z_i^{ED}
    =
    \frac{\mathbf W_{ED}\mathbf h_{\mathcal P_i}}
    {\|\mathbf W_{ED}\mathbf h_{\mathcal P_i}\|_2},
    \quad
    \mathbf z_i^{mol}
    =
    \frac{\mathbf W_{mol}\mathrm{MolEnc}(\mathcal G_i)}
    {\|\mathbf W_{mol}\mathrm{MolEnc}(\mathcal G_i)\|_2}.
\end{equation}
Here, $\mathcal G_i$ is the paired molecular input, $\mathrm{MolEnc}(\cdot)$ denotes the molecule encoder. We implement $\mathrm{MolEnc}(\cdot)$ as a lightweight molecule structure encoder
with atom-type embeddings, coordinate features, and Transformer layers, rather
than using an external pretrained molecular model. $\mathbf W_{ED}$ and $\mathbf W_{mol}$ are learnable projection matrices, and $\mathbf z_i^{ED}$ and $\mathbf z_i^{mol}$ are the $\ell_2$-normalized ED and molecule embeddings, respectively.
The similarity matrix is
\begin{equation}
    S_{ij}
    =
    (\mathbf z_i^{ED})^\top \mathbf z_j^{mol}\cdot s,
\end{equation}
where $s=\exp(\gamma)$ is a learnable logit scale. The bidirectional retrieval loss follows a symmetric contrastive objective~\citep{radford2021learning}, which is
\begin{equation}
    \mathcal L_{MER}
    =
    \frac{1}{2}
    \left[
    \mathrm{CE}(S,\mathbf y)
    +
    \mathrm{CE}(S^\top,\mathbf y)
    \right],
\end{equation}
where $\mathrm{CE}(\cdot,\cdot)$ denotes cross-entropy loss and $\mathbf y=(1,\ldots,B)$ denotes the matched ED--molecule pairs within a batch.
\paragraph{Molecule-Conditioned Electron-Density Prediction.}

For molecule-conditioned electron-density prediction, ED-DiT is used as a density-field backbone. Similar to how a distance field represents object geometry by assigning values to spatial locations, an electron-density field describes electronic structure by assigning density values to points around the molecular scaffold. This task evaluates whether the pretrained ED backbone can support density-field prediction rather than only scalar property prediction.

Given molecular coordinates and atom types, a molecule encoder extracts structural features:
\begin{equation}
    \mathbf H_{\mathrm{mol}} = \mathrm{MolEnc}(\mathcal G).
\end{equation}
For each query ED coordinate $\mathbf r$, the ED branch takes the coordinate together with a zero density feature as input, so ground-truth density values are not used as input. The pretrained ED-DiT encoder serves as a density-aware backbone, while molecular features are injected through zero-initialized cross-attention adapters inspired by ControlNet ~\citep{zhang2023adding}. The model predicts the density value at each query point:
\begin{equation}
    \hat{\rho}(\mathbf r)
    =
    g_{\phi}^{\mathrm{EDP}}
    \left(\mathbf r, \mathbf H_{\mathrm{mol}}\right).
\end{equation}
where $g_{\phi}^{\mathrm{EDP}}$ denotes the molecule-conditioned density prediction head with parameters $\phi$.
During supervised fine-tuning, the molecule encoder and adapters learn how molecular structure conditions the pretrained density-field backbone, enabling the model to recover electron density at query points without using ground-truth density as input.

\section{Experiments}

\subsection{Experimental Tasks}

We evaluate ED-DiT on EDBench using the benchmark-provided scaffold split.
The benchmark covers diverse electron-density-related tasks, including orbital energy prediction (OE), energy correction (EC), multipole moment prediction (MM), open-/closed-shell classification (OCS), molecule electron-density retrieval (MER), and molecule-conditioned electron-density prediction (EDP).

For self-supervised pretraining, we use electron-density fields from the training splits of six EDBench datasets:
\texttt{ed\_homo\_lumo\_5w},
\texttt{ed\_energy\_5w},
\texttt{ed\_multipole\_moments\_5w},
\texttt{ed\_prediction\_5w},
\texttt{ed\_open\_shell\_5w}, and
\texttt{ed\_retrievel\_5w}.
No downstream task labels are used during pretraining.

After pretraining, the learned ED-DiT representation is transferred to different downstream tasks.
For OE, EC, MM, and OCS, we evaluate both full-data and label-efficient settings using $1\%$, $10\%$, and $100\%$ of labeled training samples.
For MER and EDP, we follow the benchmark protocol and train the corresponding task-specific modules using the available labeled data.

\subsection{Evaluation Metrics}

We use task-specific metrics to evaluate different aspects of model performance.
For regression tasks, including OE, EC, MM, and EDP, we report mean absolute error (MAE) and root mean square error (RMSE).
For OCS, we report accuracy, AUROC and balanced accuracy.
For MER, we evaluate bidirectional retrieval accuracy between molecules and electron-density fields using Top-$1$ accuracy.
For all tasks, model selection is based on validation performance, and final test results are reported using the selected checkpoint.

\subsection{Baselines}
For OE, EC, MM, and OCS, we compare ED-DiT with Scratch, X-3D, and PointVector-S. Scratch uses the same architecture and electron-density point-cloud input as ED-DiT, but is trained from random initialization, allowing us to isolate the effect of self-supervised pretraining.
For MER, we compare against Geoformer \citep{wang2023geometric} with X-3D \citep{sun2024x3d} or PointVector-S \citep{deng2023pointvector} as the molecular encoder. All methods follow the official retrieval protocol, using the same test queries, one positive and ten predefined negatives per query, and the same similarity-based ranking procedure.
For the EDP task, we additionally compare with DeepDFT \citep{jorgensen2020deepdft} as a baseline.
We use the official DeepDFT implementation and adapt it to EDBench:
The model is trained on the raw \texttt{Mol1\_Dt.cube} files from the same training split and evaluate it on the same processed test points. We also compare with HGEGNN, the official EDBench baseline for molecule-conditioned electron-density prediction, reporting MAE and RMSE.

\subsection{Implementation Details}

All experiments are implemented in PyTorch and conducted on a server equipped with eight NVIDIA H100 GPUs with 80GB memory per GPU.
Data-parallel training is used for both pretraining and downstream experiments.
Unless otherwise specified, all experiments use eight data-loading workers and random seed 2026.

For pretraining, each molecule is represented by up to 2048 sampled electron-density points.
We randomly mask approximately $50\%$ of valid points and apply diffusion noise during training.
The diffusion process contains 1000 timesteps.
The ED-DiT backbone follows a Diffusion Transformer architecture with hidden dimension 256, 6 DiT blocks, and 8 attention heads. The pretraining objective combines masked denoising loss and electron-number consistency loss. The model is pretrained for 150 epochs using AdamW with learning rate $1\times10^{-4}$, weight decay 0.01, and batch size 64.

For downstream experiments, the pretrained backbone is jointly optimized with task-specific modules.
Regression tasks use prediction heads for OE, EC, and MM; OCS uses a binary classification head; MER adopts a dual-tower contrastive learning framework; and EDP uses a molecule-conditioned density prediction decoder.
All downstream models are trained for 50 epochs using AdamW with batch size 64, learning rate $1\times10^{-4}$, and weight decay 0.01.
For EDP, the backbone learning rate is reduced to $2\times10^{-5}$, and the pretrained backbone is frozen for the first 10 epochs before joint optimization.

\begin{figure}[t]
    \centering
    \includegraphics[width=0.7\linewidth]{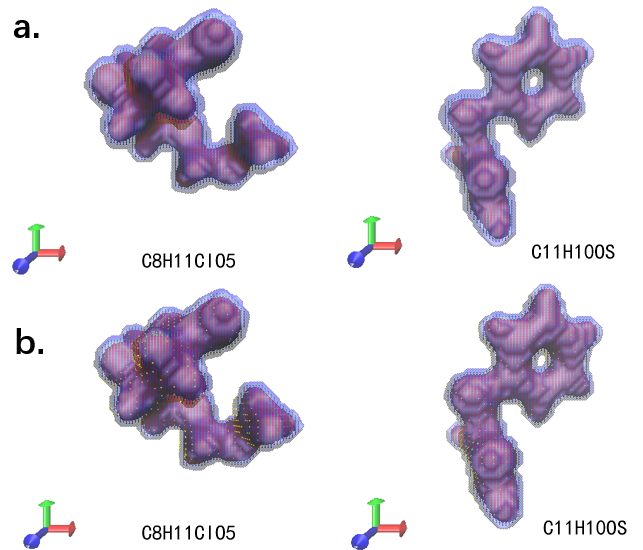}
    \caption{
    Qualitative comparison between DFT electron-density surfaces and ED-DiT predictions.
    For each molecule, (a) shows the ground-truth electron density obtained from DFT calculations. (b) shows the molecule-conditioned electron density predicted by ED-DiT and yellow points indicate locations with relatively large prediction errors.
    }
    \label{fig:edp_density_surface}
\end{figure}

\begin{figure}[t]
    \centering
    \includegraphics[width=0.8\linewidth]{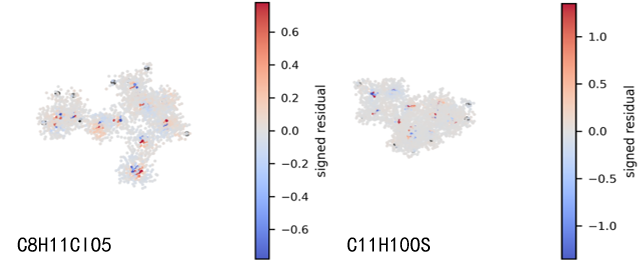}
    \caption{
    Signed residual maps for molecule-conditioned electron-density prediction.
    The residual is computed as the difference between the ED-DiT prediction and the DFT electron density.
    Red and blue denote positive and negative residuals, respectively.
    }
    \label{fig:edp_signed_residual}
\end{figure}

\begin{table}[t]
    \centering
    \caption{
    Main results on EDBench downstream tasks. Scratch denotes the same ED-DiT architecture trained from random initialization. For OE, EC, MM, and OCS, the last two columns report X-3D and PointVector-S baselines. For MER, they report Geoformer + X-3D and Geoformer + PointVector-S. E-to-M / M-to-E is P-to-G / G-to-P in Geoformer. }
    \label{tab:main_edbench_results}
    \small
    \setlength{\tabcolsep}{3.5pt}
    \renewcommand{\arraystretch}{1.1}
    \begin{tabularx}{\linewidth}{llXXXXX}
\hline
Task & Metric & Scratch & ED-DiT & X-3D / Geo. & PV-S / Geo.+PV \\
\hline
OE & RMSE $\downarrow$ & 0.0184 & \textbf{0.0080} & 0.0323 & 0.0316 \\
EC & RMSE $\downarrow$ & 43.46 & \textbf{17.93} & 250.15 & 409.22 \\
MM & RMSE $\downarrow$  & 0.5679 & \textbf{0.1722} & 1.0406 & 1.0917 \\
OCS & AUROC $\uparrow$  & 0.7193 & \textbf{0.9997} & 0.6046 & 0.5932 \\
MER$_{E\rightarrow M}$ & Top-1 $\uparrow$  & 0.9892 & \textbf{0.9978} & 0.3342 & 0.1070 \\
MER$_{M\rightarrow E}$ & Top-1 $\uparrow$  & 0.9860 & \textbf{0.9986} & 0.3806 & 0.2160 \\
\hline
\end{tabularx}

\end{table}

\begin{table}[t]
\centering
\caption{EDP results on EDBench.}
\label{tab:edp_results}
\small
\setlength{\tabcolsep}{3pt}
\begin{tabular}{llcccc}
\hline
Task & Metric & Scratch & ED-DiT & HGEGNN & DeepDFT \\
\hline
EDP & MAE $\downarrow$  & 0.4165 & \textbf{0.1213} & 0.1382 & 0.2733 \\
EDP & RMSE $\downarrow$ & 2.2474 & \textbf{1.3753} & 1.4573 & 3.2948 \\
\hline
\end{tabular}
\end{table}

\begin{table}[t]
\centering
\caption{Label-efficient fine-tuning results on property prediction and open-/closed-shell classification.}
\label{tab:label_efficiency}
\begin{tabular}{lllll}
\toprule
Task & Metric & Label & Scratch & ED-DiT \\
\midrule
OE  & RMSE $\downarrow$  & 1\%   & $0.03346$  & $\mathbf{0.02040}$ \\
OE  & RMSE $\downarrow$  & 10\%  & $0.02934$  & $\mathbf{0.01380}$ \\
OE  & RMSE $\downarrow$  & 100\% & $0.01844$  & $\mathbf{0.00801}$ \\
\midrule
EC  & RMSE $\downarrow$  & 1\%   & $1153.23$  & $\mathbf{1152.37}$ \\
EC  & RMSE $\downarrow$  & 10\%  & $184.80$   & $\mathbf{154.95}$ \\
EC  & RMSE $\downarrow$  & 100\% & $43.46$ &$\mathbf{17.9261} $\\
\midrule
MM  & RMSE $\downarrow$  & 1\%   & $1.06116$  & $\mathbf{0.69457}$ \\
MM  & RMSE $\downarrow$  & 10\%  & $0.89998$  & $\mathbf{0.38176}$ \\
MM  & RMSE $\downarrow$  & 100\% & $0.56793$  & $\mathbf{0.17216}$ \\
\midrule
OCS & AUROC $\uparrow$ & 1\%   & $0.5438$ & $\mathbf{0.9693}$ \\
OCS & AUROC $\uparrow$ & 10\%  & $0.5459$ & $\mathbf{0.9984}$ \\
OCS & AUROC $\uparrow$ & 100\% & $0.7193$ & $\mathbf{0.9997}$ \\
\bottomrule
\end{tabular}
\end{table}

\section{Main Results}
\subsection{Overall Performance on EDBench}
\paragraph{Property Prediction and Retrieval.}
We first compare ED-DiT with the same architecture trained from scratch and with available EDBench baselines. Table~\ref{tab:main_edbench_results} reports the main results on orbital energy prediction (OE), energy correction (EC), multipole moment prediction (MM), open-/closed-shell classification (OCS), and molecule--electron-density retrieval (MER). For OE, EC, and MM, lower RMSE indicates better performance. For OCS and MER, higher AUROC or Top-1 accuracy indicates better performance.

Across property prediction and classification tasks, ED-DiT consistently outperforms the scratch counterpart. It reduces RMSE from 0.0184 to 0.0080 on OE, from 43.46 to 17.93 on EC, and from 0.5679 to 0.1722 on MM, while improving OCS AUROC from 0.7193 to 0.9997. Compared with EDBench point-cloud baselines, ED-DiT achieves the best performance on all reported property and classification tasks.
For molecule--electron-density retrieval, ED-DiT achieves near-perfect Top-1 accuracy in both retrieval directions, matching or improving over the scratch model and substantially outperforming Geoformer-based baselines. These results show that electron-density pretraining benefits both molecular property prediction and cross-modal retrieval.

\paragraph{Density Prediction.}
Table~\ref{tab:edp_results} further evaluates molecule-conditioned electron-density prediction (EDP). ED-DiT achieves the best performance among all compared methods, reducing MAE from 0.4165 to 0.1213 and RMSE from 2.2474 to 1.3753 compared with the scratch counterpart. It also outperforms the official HGEGNN baseline and DeepDFT on both metrics. These results show that the pretrained electron-density backbone is effective for density-level prediction.

\paragraph{EDP Visualization.}
To complement the quantitative EDP results, we visualize molecule-conditioned electron-density prediction for two molecules in Figure~\ref{fig:edp_density_surface} and Figure~\ref{fig:edp_signed_residual}. The DFT electron-density surfaces and ED-DiT predictions show similar global morphology and local density patterns, indicating that ED-DiT can recover the spatial structure of molecular electron density from molecular conditioning. The signed residual maps show that most regions have small residuals, while larger deviations are sparse and mainly appear near local boundaries or high-variation regions. These observations suggest that the pretrained electron-density backbone provides useful spatial priors for density-field reconstruction.

\subsection{Label-Efficient Property Prediction}
To evaluate whether electron-density pretraining improves data efficiency, we further compare ED-DiT with Scratch under different fractions of labeled training data. As shown in Table~\ref{tab:label_efficiency}, ED-DiT improves over Scratch across most label settings on OE, EC, MM, and OCS. The gains are especially clear in low-label regimes: with only 1\% labels, ED-DiT substantially improves OCS AUROC from 0.5438 to 0.9693 and reduces MM RMSE from 1.06116 to 0.69457. Similar improvements are observed under the 10\% label setting, where ED-DiT achieves near-saturated OCS AUROC and notably lower MM error. These results suggest that self-supervised pretraining on electron-density fields provides transferable electronic-structure priors that are difficult to learn from limited task-specific supervision alone.

\begin{figure}[t]
    \centering
    \includegraphics[width=0.8\columnwidth]{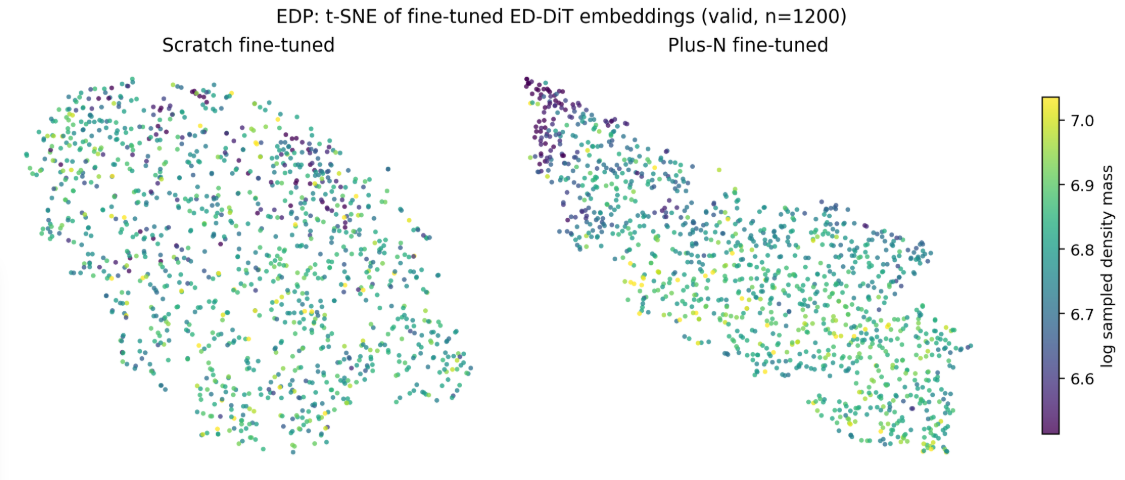}
    \caption{
    t-SNE visualization of fine-tuned ED-DiT embeddings on the EDP validation set. 
    Each point represents one molecule, and colors indicate the log-sampled density mass. 
    Pretrained fine-tuning yields a smoother organization along the density-mass signal than scratch fine-tuning.
    }
    \label{fig:tsne_edp}
    \vspace{6pt}
\end{figure}

\begin{table}[t]
\centering
\caption{Ablation of electron-number consistency on EDBench test sets.}
\label{tab:denoise_n_ablation}
{
\begin{tabular}{llcc}
\toprule
Task & Metric & Denoise & Denoise + $\mathcal{L}_N$ \\
\midrule
EDP & MAE / RMSE & 0.13106 / 1.42663 & \textbf{0.12133} / \textbf{1.37535} \\
OE & MAE / RMSE & \textbf{0.005770} / \textbf{0.007951} & 0.005842 / 0.008011 \\
EC & MAE / RMSE & 11.2720 / 18.6315 & \textbf{10.3682} / \textbf{17.9261} \\
MM & MAE / RMSE & \textbf{0.09736} / \textbf{0.15768} & 0.10925 / 0.17216 \\
OCS & Acc /BAC & 0.99200 / 0.99172 & \textbf{0.99480}  /  \textbf{0.99461} \\
MER & E-to-M / M-to-E Top-1 & 0.9974 / 0.9968   & \textbf{0.9978} / \textbf{0.9986} \\
\bottomrule
\end{tabular}
}
\end{table}

\section{Analysis}
\subsection{Representation Visualization}
To examine how electron-density pretraining shapes the representation space, we visualize fine-tuned ED-DiT embeddings on the EDP validation set using t-SNE. Each point in Figure~\ref{fig:tsne_edp} represents a molecule, and the color indicates its log-sampled density mass.

Compared with the relatively dispersed scratch embeddings, pretrained embeddings show a smoother organization, with molecules of similar density mass located closer together. This suggests that pretraining not only improves downstream performance but also encourages representations aligned with density-related physical structure.
This visualization provides qualitative evidence that the pretrained encoder captures reusable electronic-structure information. Since t-SNE is a nonlinear projection, we use it as a diagnostic visualization rather than a standalone quantitative metric. Together with the downstream results, the representation analysis supports the hypothesis that self-supervised pretraining on electron density helps learn transferable molecular representations.

\subsection{Effect of Electron-Number Consistency}
Table~\ref{tab:denoise_n_ablation} evaluates the effect of adding the electron-number consistency loss $\mathcal{L}_N$ to masked diffusion pretraining.
The constraint improves EDP, EC, and OCS, with the clearest gain on EDP. Since EDP requires recovering electron density conditioned on molecular structure, its improvement suggests that $\mathcal{L}_N$ helps the backbone learn a more physically coherent ED prior. 
 Overall, these results support using denoising plus $\mathcal{L}_N$ as the default ED-DiT objective, while treating stronger physical constraints as analysis rather than the main contribution.

\section{Conclusion}
We presented ED-DiT, a physics-guided self-supervised framework for learning transferable molecular representations from three-dimensional electron-density fields. By combining masked diffusion denoising with an electron-number consistency constraint, ED-DiT learns a reusable electronic-structure encoder that transfers across property prediction, classification, retrieval, and density prediction tasks on EDBench. The results show consistent gains over training from scratch, especially under limited supervision, and suggest that electron density provides an effective pretraining signal for molecular representation learning.

\section{Supplementary Material}

\subsection{Electron-Number Consistency Substantially Improves Physical Consistency}

\begin{figure}[!t]
    \centering
    \includegraphics[width=\columnwidth]
    {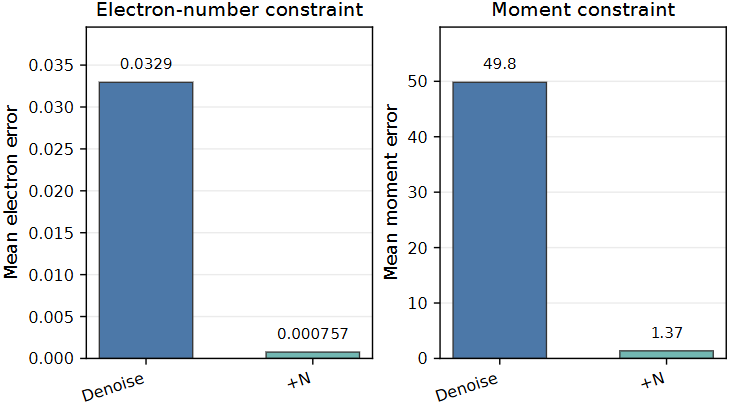}
    \caption{
    Physical-consistency diagnostics for denoising-only pretraining and its variant with electron-number consistency (\(N\)). 
    Adding \(N\) substantially reduces both the mean electron-number error and the mean moment error, demonstrating improved physical consistency.
    }
    \label{fig:supp_denoise_plusN_physical_consistency}
\end{figure}

We further investigate the role of electron-number consistency (\(N\)) in physics-guided pretraining by comparing the denoising-only objective with its variant augmented by the electron-number constraint. As shown in Figure~\ref{fig:supp_denoise_plusN_physical_consistency}, electron-number consistency substantially improves physical consistency: the mean electron-number error is reduced from \(3.29\times 10^{-2}\) to \(7.57\times 10^{-4}\), while the mean moment error decreases from \(49.8\) to \(1.37\). These large reductions indicate that \(N\) provides a strong global physical constraint that guides the model to preserve conserved electron-density quantities, rather than merely optimizing pointwise denoising accuracy.

\subsection{Selection of the Electron-Number Loss Weight}

\begin{figure}[!t]
    \centering
    \includegraphics[width=\columnwidth]
    {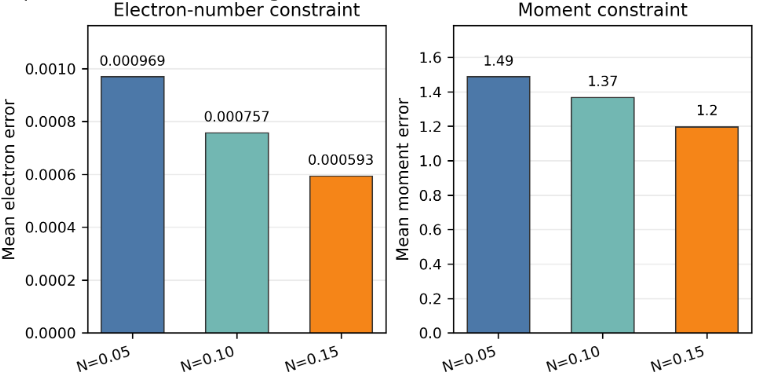}
    \caption{
    Physical-consistency diagnostics under different electron-number loss weights \(\lambda_N\). 
    }
    \label{fig:supp_lambda_n_physical_consistency}
\end{figure}

We further study the choice of the electron-number loss weight \(\lambda_N\) by comparing three settings: \(0.05\), \(0.10\), and \(0.15\), denoted as N005, N010, and N015, respectively. As shown in Figure~\ref{fig:supp_lambda_n_physical_consistency}, increasing \(\lambda_N\) consistently improves physical consistency: the mean electron-number error decreases from \(9.69\times 10^{-4}\) to \(7.57\times 10^{-4}\) and further to \(5.93\times 10^{-4}\), while the mean moment error decreases from \(1.49\) to \(1.37\) and further to \(1.20\).

However, stronger physical regularization does not necessarily lead to better downstream transfer performance. As shown in Table~\ref{tab:lambda_n_sweep}, N010 achieves the best performance on three out of five tasks, while its performance on the remaining two tasks is also only marginally worse. N010 provides a better overall balance across physical consistency and downstream transfer. Therefore, we choose N010 as the default electron-number loss weight in our full-spec physics-guided pretraining setting.

\begin{table}[!t]
\centering
\caption{
Downstream performance under different electron-number loss weights \(\lambda_N\). 
}
\label{tab:lambda_n_sweep}
\small
\setlength{\tabcolsep}{6pt}
\renewcommand{\arraystretch}{1.08}
\begin{tabular}{@{}lcccc@{}}
\toprule
Task & Metric & N005 & N010 & N015 \\
\midrule
EDP & MAE
& 0.12339
& \textbf{0.12133}
& 0.12757 \\

OE & MAE
& 0.005847
& \textbf{0.005842}
& 0.005965 \\

EC & MAE
& \textbf{9.8218}
& 10.3682
& 11.0886 \\

MM & MAE
& \textbf{0.09589}
& 0.10925
& 0.10972 \\

OCS & AUROC
& 0.99965
& \textbf{0.99973}
& \textbf{0.99973} \\
\bottomrule
\end{tabular}
\end{table}

\subsection{Qualitative Electron-Density Prediction Examples}

\begin{figure*}[!t]
    \centering

    \begin{minipage}[t]{0.485\textwidth}
        \vspace{0pt}
        \centering
        \includegraphics[width=\linewidth]
        {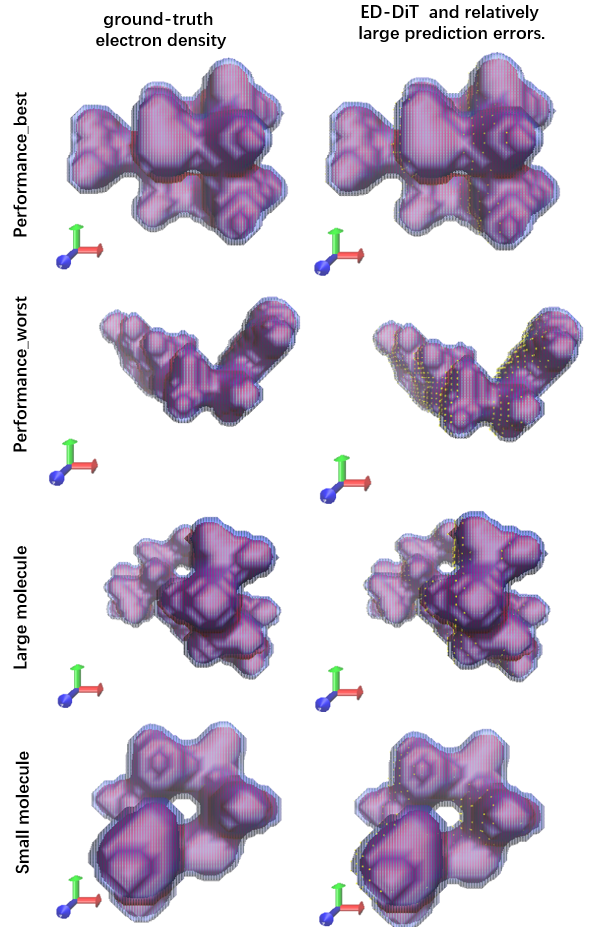}
        \caption{
        Qualitative comparison between DFT electron-density surfaces and ED-DiT predictions across different molecule-level conditions. 
        For each molecule, the left panel shows the ground-truth electron density obtained from DFT calculations, while the right panel shows the molecule-conditioned electron density predicted by ED-DiT. 
        Yellow points indicate spatial locations with relatively large prediction errors.
        }
        \label{fig:supp_ed_density_case_study}
    \end{minipage}
    \hfill
    \begin{minipage}[t]{0.485\textwidth}
        \vspace{0pt}
        \centering
        \includegraphics[width=\linewidth]
        {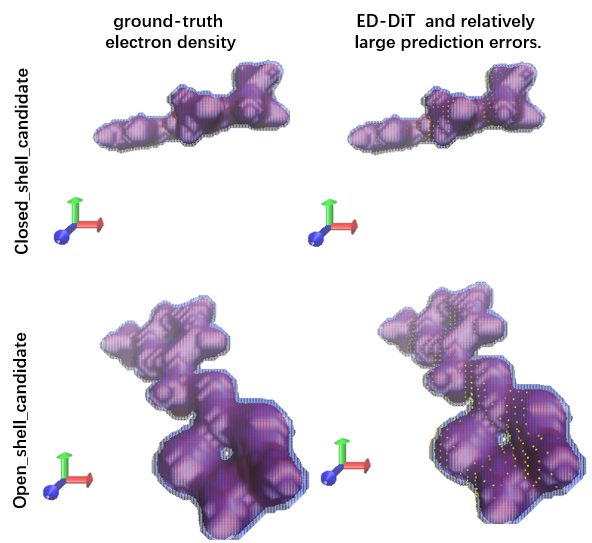}
        \caption{
        Qualitative electron-density prediction examples for
        closed-shell and open-shell candidates. The visualization
        convention follows Figure~\ref{fig:supp_ed_density_case_study}.
        }
        \label{fig:supp_ed_density_shell_case_study}
    \end{minipage}

\end{figure*}

To qualitatively assess molecule-conditioned electron-density prediction, we compare DFT electron-density surfaces with the corresponding ED-DiT predictions. Figure~\ref{fig:supp_ed_density_case_study} shows representative examples across different molecule-level conditions, including the best-performing case, the worst-performing case, a large molecule, and a small molecule. Figure~\ref{fig:supp_ed_density_shell_case_study} further shows examples for closed-shell and open-shell candidates. For each molecule, the left panel shows the ground-truth electron density obtained from DFT calculations, while the right panel shows the molecule-conditioned electron density predicted by ED-DiT. Yellow points indicate spatial locations with relatively large prediction errors.

\subsection{Error Analysis under Local Electron-Density Variation}

In the main paper, the signed residual visualizations suggest
that larger prediction errors tend to occur in regions with
strong local electron-density variation. To 
validate this qualitative observation, we analyze the prediction
error under different levels of ground-truth local density
variation.

For each query point $i$, we compute its local density variation
using its $k$ nearest neighbors:
\begin{equation}
V_i =
\frac{1}{|\mathcal{N}_k(i)|}
\sum_{j\in\mathcal{N}_k(i)}
\frac{|\rho_j-\rho_i|}
{\|\mathbf{r}_j-\mathbf{r}_i\|_2+\epsilon},
\end{equation}
where $\rho_i$ and $\mathbf{r}_i$ denote the ground-truth
electron density and spatial coordinate, respectively. We use
$k=8$ and exclude the query point itself from its neighborhood.

For each molecule, valid query points are ranked according to
$V_i$ and divided into four equally sized quartiles. Q1 contains
the points with the lowest local density variation, whereas Q4
contains those with the highest local density variation. MAE and
RMSE are first computed for each molecule and quartile and are
then macro-averaged across test molecules.

\begin{table*}[!t]
\centering
\setlength{\belowcaptionskip}{6pt}
\caption{
Electron-density prediction errors under different levels
of local ground-truth density variation. Q1 contains the smoothest
regions, whereas Q4 contains the regions with the highest local
variation. All values are macro-averaged over test molecules.
The RMSE improvement is calculated relative to the scratch model.
}
\label{tab:local_variation_results}
\small
\setlength{\tabcolsep}{9pt}
\renewcommand{\arraystretch}{1.12}
\begin{tabular}{@{}lccccc@{}}
\toprule
& \multicolumn{2}{c}{MAE $\downarrow$}
& \multicolumn{2}{c}{RMSE $\downarrow$}
& \multirow{2}{*}{\shortstack{RMSE improvement\\$\uparrow$}} \\
\cmidrule(lr){2-3}
\cmidrule(lr){4-5}
Local variation
& Scratch
& ED-DiT
& Scratch
& ED-DiT
& \\
\midrule
Q1 (Smoothest)
& 0.2513
& \textbf{0.0194}
& 0.2624
& \textbf{0.0262}
& 90.0\% \\

Q2
& 0.2321
& \textbf{0.0203}
& 0.2442
& \textbf{0.0272}
& 88.8\% \\

Q3
& 0.1891
& \textbf{0.0274}
& 0.2099
& \textbf{0.0382}
& 81.8\% \\

Q4 (Highest variation)
& 0.9753
& \textbf{0.4182}
& 4.1111
& \textbf{2.0080}
& 51.2\% \\
\bottomrule
\end{tabular}
\end{table*}

\begin{figure}[!t]
\centering
\includegraphics[width=0.60\columnwidth]
{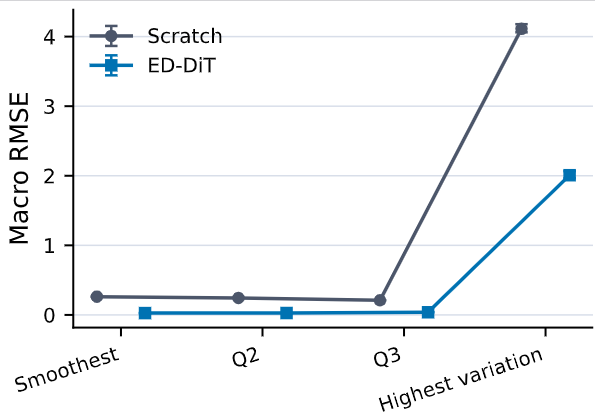}
\caption{
Quantitative validation of the relationship between local electron-density variation and prediction error. Query points are divided into quartiles within each molecule according to their ground-truth local density variation. Errors remain relatively low in Q1-Q3 but increase sharply in Q4, supporting the qualitative observation in the main paper that larger residuals are concentrated in high-variation regions. Error bars denote molecule-level bootstrap 95\% confidence intervals.
}
\label{fig:local_variation_rmse}
\end{figure}

\begin{table}[!t]
\centering
\caption{
Long-tail diagnosis of absolute prediction errors in the
Q4 highest-variation regions.
}
\label{tab:q4_tail_analysis}
\small
\setlength{\tabcolsep}{8pt}
\renewcommand{\arraystretch}{1.10}
\begin{tabular}{@{}lcc@{}}
\toprule
Metric & Scratch & ED-DiT \\
\midrule
Median absolute error
& 0.1472
& \textbf{0.0527} \\

P99 absolute error
& 17.2010
& \textbf{7.1947} \\

P99.9 absolute error
& 54.4980
& \textbf{28.0745} \\

Maximum absolute error
& 2461.9356
& \textbf{2415.3700} \\

Q4 MAE
& 0.9753
& \textbf{0.4182} \\

Q4 RMSE
& 4.1111
& \textbf{2.0080} \\
\bottomrule
\end{tabular}
\end{table}

Table~\ref{tab:local_variation_results} and Figure~\ref{fig:local_variation_rmse} show that prediction errors remain low across Q1–Q3 but rise sharply in Q4. This aligns with the main paper's qualitative observation that larger residuals concentrate in regions of strong local electron-density variation. ED-DiT outperforms the scratch model across all quartiles. In the most challenging Q4 regions, ED-DiT lowers MAE from 0.9753 to 0.4182 and RMSE from 4.1111 to 2.0080, a relative RMSE reduction of 51.2\%. These results confirm that pretraining benefits density prediction even under the strongest local variation.

Table~\ref{tab:q4_tail_analysis} confirms that the high Q4 RMSE reflects a genuine long-tail problem, not artifacts of numerical instability or grouping decisions. For ED-DiT, the median absolute error in Q4 is just 0.0527, but the P99 and P99.9 errors jump to 7.1947 and 28.0745, which means that squared error is dominated by a tiny fraction of extreme high-density points. ED-DiT consistently beats the scratch model on both typical and tail errors, though the model's performance on the most extreme densities leaves room for improvement.

\subsection{Discussion and Future Work}

Overall, the results suggest that electron-density reconstruction provides a transferable pretraining signal across property prediction, classification, retrieval, and density prediction. The consistent gains over training from scratch, especially under limited supervision, indicate that ED-DiT captures reusable electronic-structure priors. The local-variation analysis further shows that pretraining improves density prediction across both smooth and rapidly varying regions, although errors remain concentrated in a small number of extreme high-density points.

Future work will focus on improving prediction in these challenging regions through incorporating more physical constraints, multi-scale feature extraction, and objectives that better account for rare high-density points. It will also be important to evaluate transfer across broader chemical domains, different electronic-structure settings, higher spatial resolutions, and larger molecular systems. To support denser electron-density sampling, sparse, local, or hierarchical attention could be explored to reduce the computational cost of global self-attention.
\bibliographystyle{unsrtnat}
\bibliography{references}

\end{document}